\documentclass[journal]{IEEEtran}

\usepackage{graphicx}
\usepackage{wrapfig}
\usepackage{amsmath}
\usepackage{caption,subcaption,graphicx}

\usepackage{esvect}
\usepackage[utf8]{inputenc}
\usepackage{color}
\usepackage[square,sort,comma,numbers]{natbib}
\usepackage{amssymb}

\usepackage[numbers]{natbib}
\usepackage{empheq} 
\usepackage{amsmath}
\usepackage{commath}
\usepackage{indentfirst}
\usepackage{amsfonts}
\usepackage{xcolor}

\newcommand{\etc}{etc}

\newcommand{\eg}{e.g.}

\usepackage[shortlabels]{enumitem}
\usepackage{epstopdf}

   \usepackage{tabu}
\usepackage[colorlinks = true, linkcolor = green, urlcolor  = black, citecolor = blue, anchorcolor = blue]{hyperref}

\allowdisplaybreaks[3]

\begin{document}

\title{Physics-Consistent Data-driven Waveform Inversion with Adaptive Data Augmentation}

\author{Renán Rojas-Gómez$^{\dagger, \diamond}$, Jihyun Yang$^{\dagger, \#}$, Youzuo Lin$^{\dagger, \star}$, James Theiler$^{\dagger}$, and Brendt Wohlberg$^{\dagger}$
\vspace{\baselineskip}

$\dagger$:~Los Alamos National Laboratory\\
$\diamond$:~Department of Electrical and Computer Engineering, University of Illinois at Urbana-Champaign\\
		  $\#$:~Department of Geophysics, Colorado School of Mines\\
\thanks{$\star$~Correspondence to: Y. Lin, ylin@lanl.gov.}
}

\markboth{IEEE GEOSCIENCE AND REMOTE SENSING LETTERS}%
{Shell \MakeLowercase{\textit{et al.}}: Bare Demo of IEEEtran.cls for Journals}

\maketitle


\begin{abstract}
Seismic full-waveform inversion~(FWI) is a nonlinear computational imaging technique that can provide detailed estimates of subsurface geophysical properties. Solving the FWI problem can be challenging due to its ill-posedness and high computational cost. In this work, we develop a new hybrid computational approach to solve FWI that combines physics-based models with data-driven methodologies. In particular, we develop a data augmentation strategy that can not only improve the representativity of the training set, but also incorporate important governing physics into the training process and therefore improve the inversion accuracy. To validate the performance, we apply our method to synthetic elastic seismic waveform data generated from a subsurface geologic model built on a carbon sequestration site at Kimberlina, California. We compare our physics-consistent data-driven inversion method to both purely physics-based and purely data-driven approaches and observe that our method yields higher accuracy and greater generalization ability. 

\end{abstract}


\IEEEpeerreviewmaketitle

\section{Introduction}


\IEEEPARstart{I}n solid earth geosciences, characterizing the subsurface geology is crucial for energy exploration, civil infrastructure, groundwater contamination and remediation, \etc. However, nearly all of the earth’s interior is inaccessible to direct observation. Inference of unknown subsurface properties therefore relies on indirect and limited geophysical measurements taken at or near the surface. Seismic inversion attempts to reconstruct an image of subsurface structures from measurements of natural or artificially produced seismic waves that have travelled through the subsurface. A \textit{forward model} describes how the observations depend on the subsurface map, while the inverse problem involves inferring that map from the observations. The forward model of wave propagation is nonlinear.
Travel-time inversion methods~\cite{tarantola-2005-inverse} are based on a linear approximation of the forward model,
while seismic full-waveform inversion~(FWI) addresses the full non-linear problem, leading to superior inversion accuracy and resolution~\cite{introduction-2014-Virieux}. 

The FWI problem is challenging due to the non-linearity of the forward model and its under-determined nature.
Conventional computational methods for solving FWI are based on optimization techniques and generic regularization~\cite{introduction-2014-Virieux}. For simplicity of description, we call these approaches ``physics-based FWI methods'' to distinguish them from data-driven methods, and from our proposed hybrid approach. The major advantage  of  these  methods  is  their  robustness  to  out-of-distribution  data  due  to  noise,  change  of  station,  and  other external factors, while the main disadvantage is their computational expense. The primary computational  cost  involved  in  FWI is associated with the solution of the wave equation, and is affected by the details of the finite difference solver, the velocity model dimension, sources and receivers.

Most existing regularization techniques used for solving FWI employ generic functions, such as $\ell_1$ or $\ell_2$ penalties on the gradient of the solution~\cite{lin-2015-acoustic}. 

Recently, a new class of algorithm has been developed, based on machine learning applied to large datasets that are produced from many runs of the forward physics model. In direct end-to-end learning~\cite{Yang-2019-Deep, Araya-2018-Deep, Tomography-2018-Farris}, a large number of velocity maps and corresponding seismic waveforms (usually constructed through extensive simulation) are used as training data in learning the mapping from seismic waveform to velocity map. In low-wave number learning~\cite{Ovcharenko-2019-Deep, Progressive-2019-Hu}, this type of learning approach is used to predict an initial velocity map with low-frequency component, which is then used as the initial guess for traditional physics-based optimization. 

Here, we describe an approach for seismic FWI that incorporates the physics model into the learning procedure. Specifically, this physics-consistent data-driven full waveform inversion consists of a carefully designed encoder-decoder-structured neural network and an adaptive data augmentation technique. This augmentation employs the forward model to produce new training data that are more representative of the solution we seek. To validate its performance, we applied our inversion method to detect carbon sequestration leakage using synthetic seismic data sets generated using a subsurface model for a potential CO$_2$ storage site at Kimberlina, California~\citep{Downhole-2019-Buscheck}.

\section{Background}
\label{sec:bk}

\subsection{Governing Physics: the Forward Model}

Mathematically, the forward model can be expressed in terms of the seismic elastic-wave partial differential equation~\cite{introduction-2014-Virieux}:

\begin{align}
    \nonumber
    \rho({r}) \frac{\partial^2 u({r}, t)}{ \partial t^2} = 
    & (\lambda({r}) + \mu({r}))  \nabla (\nabla \cdot u({r}, t)) \\
    & + \mu({r}) \nabla^2 u({r}, t)  + s({r},\, t),
    \label{eq:ForwardElastic}
\end{align}
where $\rho ({r})$ is the density at spatial location ${r}$, $\lambda({r})$ and $\mu({r})$ are the Lam\'e parameters, $s({r},\, t)$ is the source term, $u(\mathbf{r}, t)$ is the displacement wavefield, $t$ represents time, and $\cdot$ is the divergence operator. When fluid such as supercritical CO$_2$ leaks into the subsurface formation, the geophysical parameters of P-wave and S-wave velocities will be changed correspondingly.

Instead of inverting for $\rho ({r})$, $\lambda({r})$ and $\mu({r})$, it is customary to invert for a velocity map $m \in \mathbb{R}^{M \times N}$, where $M$ and $N$ are its vertical and lateral dimensions, respectively; here $m$ refers to either P-wave or S-wave velocity can be expressed as a function of $\rho ({r})$, $\lambda({r})$ and $\mu({r})$. Similarly, we denote a seismic data observation $d_{\text{obs}}\in \mathbb{R}^{T \times S \times R}$, where $T$ corresponds to the number of samples in the temporal domain, $S$ to the number of sources and $R$ to the number of receivers used in the data acquisition process. The seismic data can be expressed in terms of a highly nonlinear forward mapping $f$:
 \begin{equation}
    d_{\text{obs}} = f({m}).
    \label{eq:forward}
 \end{equation}


\subsection{Physics-Based Full-waveform Inversion}

Various explicit regularization techniques have been developed to stabilize the computation of seismic inversion, including $\ell_1$-norm~\cite{lin-2015-acoustic} or $\ell_2$-norm~\cite{fichtner2010full} methods. Given the forward model in Eq.~\eqref{eq:forward}, the regularized seismic FWI can be posed as
\begin{equation}
m = \underset{{m}}{\operatorname{argmin}} \left 
\{\left \| d - f({m})\right \| _2 ^2 + \lambda\, R({m}) 
\right \},
\label{eq:MisFit}
\end{equation}
where ${d}$ represents a recorded/field waveform dataset, 
$f({m})$ is the corresponding forward modeling result, $ \left \| d - f({m})\right \| _2 ^2$ is the data misfit, $||\cdot ||_2$ stands for the $\ell_2$ norm, $\lambda$ is a regularization parameter and $R({m})$ is the regularization term. Note that Eq.~\eqref{eq:MisFit} is a general formulation for regularized FWI. More effective regularization techniques have been developed for time-lapse monitoring~\cite{Maharramov-2016-Time}.

\section{Physics-Consistent Data-Driven Full-waveform Inversion}
\label{sec:PIDD}

\subsection{Data-Driven Inversion and Network Structure}
\label{sec_network}

 A data-driven FWI structure based on an \textit{encoder-decoder} architecture~\cite{wu2018inversionnet}, denoted $\mathcal{G}$ and characterized by hyperparameters $\boldsymbol{\theta}$, is proposed to approximate the inverse mapping $f^{-1}$ and obtain accurate velocity map predictions $\hat{m}(\boldsymbol{\theta})\triangleq \mathcal{G}(\boldsymbol{\theta}, d_{\text{obs}})$ under a supervised learning scheme. Optimal parameters $\boldsymbol{\theta}^{*}$ are obtained by adapting the architecture to a representative training set with $L$ samples $\{d_{\text{obs},\ell},m_{\ell}^{*}\},\ell\in\{0,L-1\}$.

We  choose the mean-absolute error (MAE) as our optimality criterion:
 \begin{align}
    \label{eq_mae}
    \boldsymbol{\theta}^{*}=\underset{\boldsymbol{\theta}}{\text{argmin}}\ \frac{1}{L}\sum_{\ell=0}^{L-1}\|m_{\ell}^{*}- \hat{m}_{\ell}(\boldsymbol{\theta})\|_{1}.
 \end{align}
For a more detailed discussion of loss function selection, please refer to our earlier work \cite{wu2018inversionnet}. 


Our data-driven inversion network structure consists of an \textit{encoder} network and a \textit{decoder} network~\cite{wu2018inversionnet}. Full details of our model are provided in the Supplementary Material.  

\subsection{Data Description}
\label{sec_data}
We apply our method to detect CO$_2$ leakage in subsurface. To the best of our knowledge, there is no real seismic data related to our problem of interest. We use the simulated Kimberlina dataset from Lawrence Livermore National Laboratory~(and that we refer to here as \texttt{CO$_{2}$leak}). The aim of the Kimberlina dataset is to understand and assess the effectiveness of various geophysical monitoring techniques in detecting CO$_{2}$ shallow leakage in the wellbore~\cite{Characterizing-2017-DOE}. A portion of the of Kimberlina Simulations can be be downloaded from the DOE-EDX platform~\cite{NETL-2018-Kimberlina}. The Kimberlina dataset is generated based on a hypothetical numerical model built on the geologic structure of a commercial-scale geologic carbon sequestration (GCS) reservoir at the Kimberlina site in the southern San Joaquin Basin, 30 km northwest of Bakersfield, CA, USA. The P-wave and S-wave velocity maps used in this work belong to the geophysical model, which is created based on the realistic geologic-layer properties from the GCS site~\cite{Downhole-2019-Buscheck}.


The \texttt{CO$_{2}$leak} dataset contains 991 CO$_{2}$ leakage scenarios, each simulated over a duration of 200 years, with 20 leakage maps provided (ie, at every ten years) for each scenario. We obtain synthetic seismograms from elastic forward modeling on \texttt{CO$_{2}$leak} velocity maps. First, one-second traces with a time interval of $0.5$ms using 7 sources and 114 receivers are generated. We then down-sample each trace by a factor of $2$, resulting in a temporal dimension of $1000$ time steps. The sources and receivers are evenly distributed along the top of the model, with depths of 5m and 20m, respectively. The source interval is 125m, and the receiver interval is 15m. We use a Ricker wavelet with a central frequency of $25$Hz as the source to generate simulated seismic waves due to its empirical success in processing seismic field data~\cite{fichtner2010full}. 
The synthetic data is the staggered grid solution of the elastic wave equation using a finite-difference scheme with a perfectly matched layered absorbing boundary condition. 

\subsection{Data Augmentation: Incorporation of Physics Knowledge}
\label{sec_augmentation}
The \texttt{CO$_{2}$leak} dataset includes $19,600$ velocity maps of $141 \times 341$ grid points describing CO$_{2}$ and brine leakage plumes evolving with time. It is of practical interest to detect plumes of leaking CO$_{2}$ while they are still small.
This is particularly challenging when the available training data is dominated by large plumes.
Thus, \texttt{CO$_{2}$leak} presents the opportunity to evaluate the \textit{generalization} of data-driven inversion with respect to different plume sizes.

Along with the data pairs included in the dataset, the ground-truth CO$_{2}$ and brine mass information for each sample, is provided. Based on this, the full dataset is split into four parts, according to their CO$_{2}$ leak mass plus brine leak mass: \textit{tiny} plumes (from $3.53\times 10^{2}$ to $9.10\times 10^{6}$Kg), \textit{small} plumes (from $9.10\times 10^{6}$ to $2.67\times 10^{7}$Kg), \textit{medium} plumes (from $2.67\times 10^{7}$ to $8.05\times 10^{7}$Kg), and \textit{large} plumes (from $8.05\times 10^{7}$ to $1.62\times 10^{9}$ Kg). These cover $20\%$, $20\%$, $20\%$, and $40\%$ of the data samples, respectively. Figure~5 in the Supplementary Material shows representative labels from each subset.

While conventional data augmentation techniques (such as rotation, flip, scale etc.) have proved to be effective for image processing applications, it is not clear that they have a useful role to play in our application. Our adaptive data augmentation scheme provides additional training data that is not only physically meaningful but also more closely related to the target unlabelled data that we are trying to invert. We summarize our augmentation method as the following four steps (illustrated in Fig.~\ref{fig_distsampling}; a detailed description is provided in the Supplementary Material):
\begin{enumerate}[i.]
\item Estimate approximate solver $\mathcal{G}(\boldsymbol{\hat{\theta}}, d_{\text{obs}})$;

\item Generate approximate velocity maps from unlabeled data $\hat{m}_{r}=\mathcal{G}(\boldsymbol{\hat{\theta}}, d_{\text{obs},r})$;

\item Create seismic data using forward model $\hat{d}_{\text{obs}, r}=f(\hat{m}_{r})$;

\item Add new pairs to the original training set.
\end{enumerate}
The augmented dataset plays a key role in model accuracy because it will not only carry useful physics information, but also provides examples of velocity maps that are consistent with the target geology feature of interests. Furthermore, the full augmentation process can be applied in an iterative fashion by re-training the approximate solver $\mathcal{G}(\boldsymbol{\hat{\theta}}, d_{\text{obs}})$ based on the extended training set in order to generate new approximate velocity maps $\hat{m}_{r}$. This approach allows further refinement of the mapping between velocity and seismic subdomains, as empirically shown in Section \ref{sec_transferability}.

\begin{figure*}[ht]
    \centering
    \includegraphics[width=0.55\textwidth]{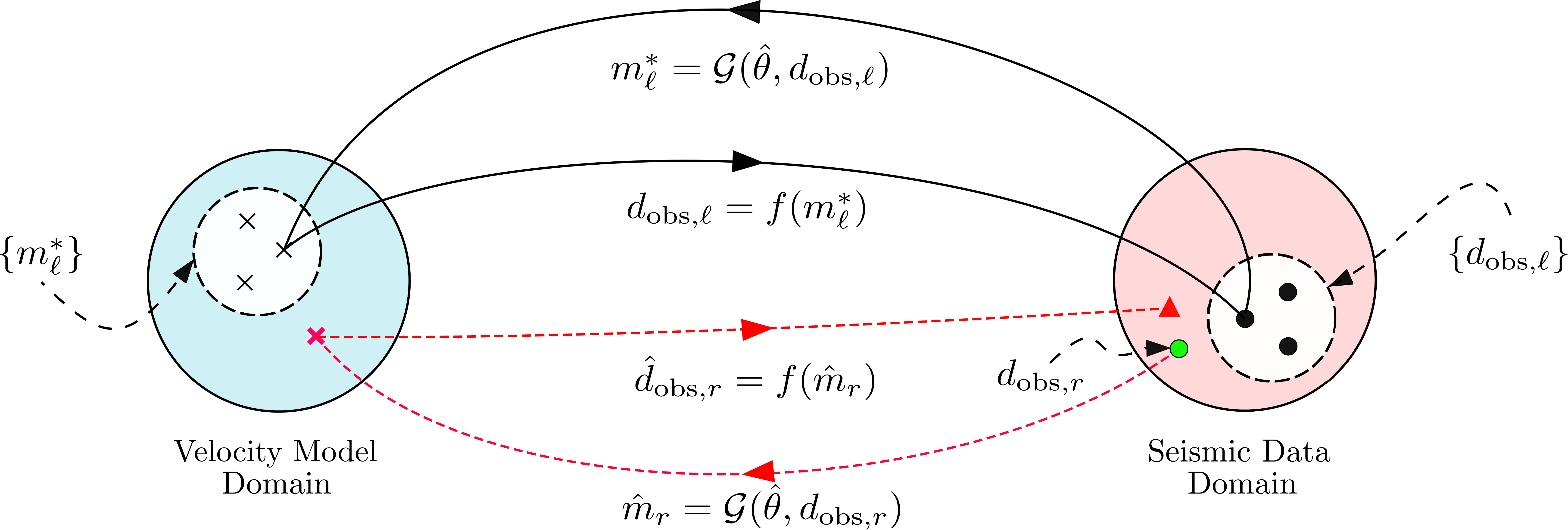}
    \caption{Adaptive Data Augmentation: Approximate solver $\mathcal{G}(\hat{\theta},d_{\text{obs}})$ is fully-trained over labeled set $\{m^{*}_{\ell},d_{\text{obs}, \ell}\}$, and applied to unlabeled seismic data $d_{\text{obs},r}$ to generate new velocity maps $\hat{m}_{r}$. Physically-coherent seismic data $\hat{d}_{\text{obs}, r}$ is then generated using the forward model $f$, producing a new labeled set $\{\hat{m}_{r}, \hat{d}_{\text{obs}, r}\}$ which is added to the original training set.}
    \label{fig_distsampling}
\end{figure*}

\section{Numerical Experiments}
\label{sec:Results}
\subsection{Experimental Setup}
For all evaluations, the training process is performed using a fixed stopping criterion of $250$ epochs, random weight initialization, an initial learning rate of $10^{-3}$ and its subsequent adaptive optimization via ADAM \cite{kingma_2014_adam}. We found that $250$ epochs is appropriate to guarantee diversity in the reconstructed velocity maps while avoiding overfitting. The training is performed using fixed batch sizes of $50$, where both features and labels are normalized prior to their use in the training and inference processes. We created $392$ batches in total. In consequence, reported loss values are computed based on normalized training and testing pairs. All training and testing routines are implemented using \textit{PyTorch} code and executed on four \textit{NVIDIA GeForce GTX 1080} GPUs. We  mention that there are various inversion strategies for elastic FWI~\cite{Strategies-2014-Raknes}, and we provide the inversion of P-wave velocities in all tests considering its higher relative sensitivity to the response of CO2 leakage. The S-wave velocities can be obtained through either sequential inversion or empirical relations~\cite{Strategies-2014-Raknes}.

\subsection{Generalization Performance without Data Augmentation}
We assess network generalization performance by first training over one subset of the training data (\eg, \textit{large} plumes) and testing on a different subset (\eg, \textit{tiny} plumes). For each case, we quantify how informative is our data augmentation approach by measuring the Mean Absolute Error (MAE), as expressed in Eq. \ref{eq_mae}, attained by the network along epochs. We show the reconstruction accuracy along epochs $\varepsilon(\boldsymbol{\theta}_{i})$ as the MAE normalized with respect to the velocity map dimensions to clearly depict our method's effect in the predictions:
 \begin{align}
    \label{eq_maelog}
    \varepsilon(\boldsymbol{\theta}_{i})=& \frac{1}{\hat{L}MN}\sum_{\ell=0}^{\hat{L}-1}\|m_{\ell}^{*}- \hat{m}_{\ell}(\boldsymbol{\theta}_{i})\|_{1},
 \end{align}
where $\boldsymbol{\theta}_{i}$ corresponds to the set of hyperparameters at the $i^{\text{th}}$ epoch, and $\hat{L}$ corresponds to the size of the validation set.

\begin{figure}[t]
  \begin{subfigure}[b]{1\linewidth}
    \centering
    \includegraphics[width=0.8\textwidth]{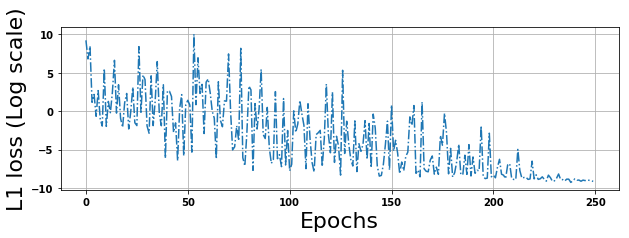}
    \phantomsubcaption\label{fig_prelim01a}
    \centerline{\small (a) Testing loss curve.}
      \vspace{0.5\baselineskip}
  \end{subfigure}

    \centering
  \begin{subfigure}[b]{0.18\textwidth}
      \centering
      \centerline{\includegraphics[width=1\textwidth]{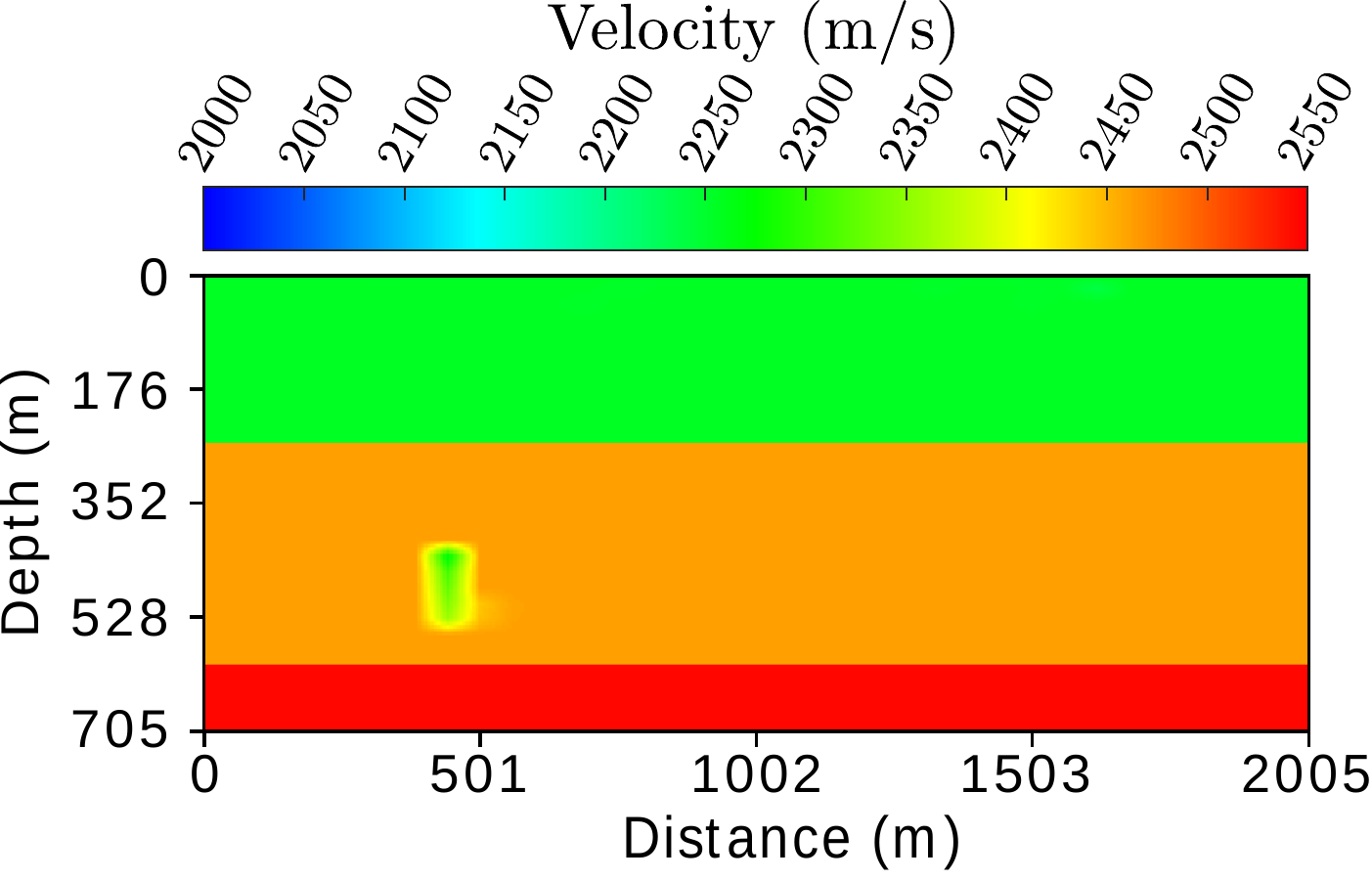}
      \phantomsubcaption\label{fig_prelim01b}}
      \centerline{\small (b) \textit{Tiny} label.}
  \end{subfigure}
    \begin{subfigure}[b]{0.18\textwidth}
      \centering
      \centerline{\includegraphics[width=1\textwidth]{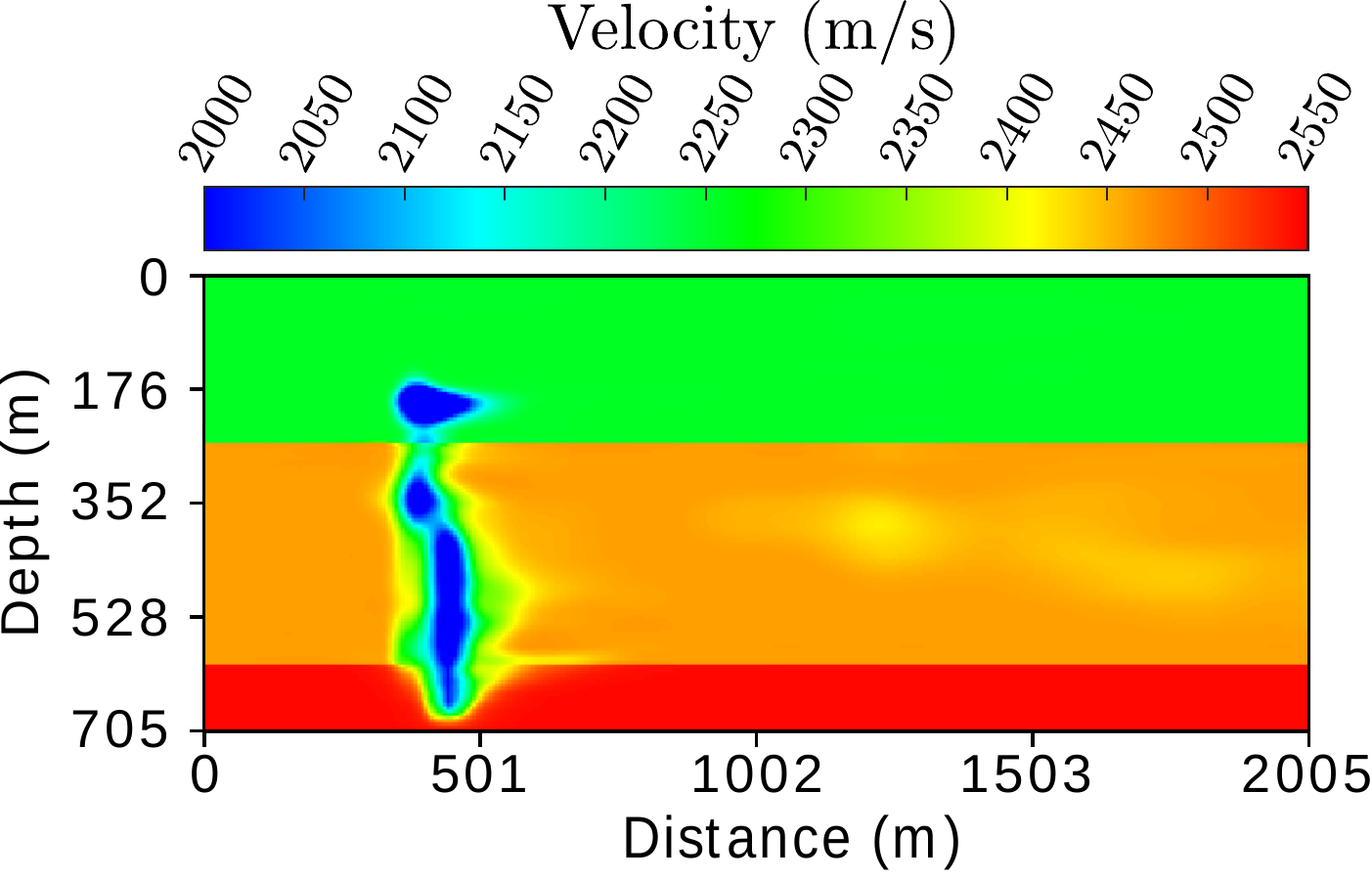}
      \phantomsubcaption\label{fig_prelim01c}}
      \centerline{\small (c) \textit{Tiny} prediction.}
  \end{subfigure}
  \caption{Transferability: Training on the \textit{large} subset, testing on the \textit{tiny} subset (0.205~(MAE) and 0.899~(SSIM)).}
  \label{fig_prelim01}
  \end{figure}

Figure~\ref{fig_prelim01a} shows the testing loss curve at each epoch. Although decreasing, the curve shows that the network is not able to accurately reconstruct the velocity maps under this scenario: given the reduced plume size in the testing set, an $\ell_{1}$ loss value of $-8.7$ indicates either a large intensity deviation among ground-truth and prediction, or a very low intersection over the union between them. This is reflected in Figures~\ref{fig_prelim01b} and \ref{fig_prelim01c}, which show a \textit{tiny} ground-truth velocity map and its prediction by the network trained over \textit{large} plumes. In general, the network tries to generate plumes of large size, which is consistent with the plumes on which it was trained.

    \begin{figure}[h]
    \begin{minipage}[b]{1\linewidth}
    \centering
    \centerline{\includegraphics[width=0.8\textwidth]{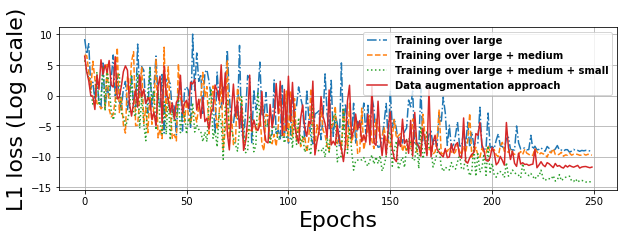}}
    \end{minipage}
    \caption{Reconstruction results for our data Augmentation approach applied to labeled data: Testing MAE loss value for each training epoch.}
    \label{fig_loss_test03}
  \end{figure}

  \begin{figure}[h]
    \begin{minipage}[b]{1\linewidth}
    \centering
    \centerline{\includegraphics[width=0.8\textwidth]{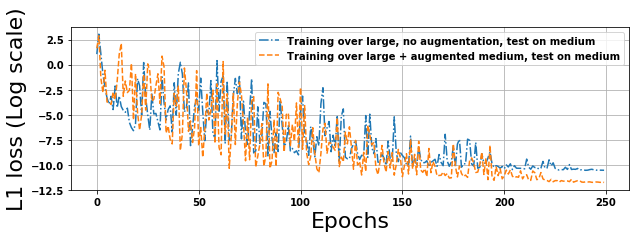}}
    \end{minipage}
    \caption{Reconstruction results for our data Augmentation approach applied to unlabeled data: Testing MAE loss value for each training epoch.}
    \label{fig_loss_test04}
  \end{figure}
  
  \begin{figure*}[t]
    \centerline{
    \begin{minipage}[b]{0.195\textwidth}
      \centering
      \centerline{\includegraphics[width=1\textwidth]{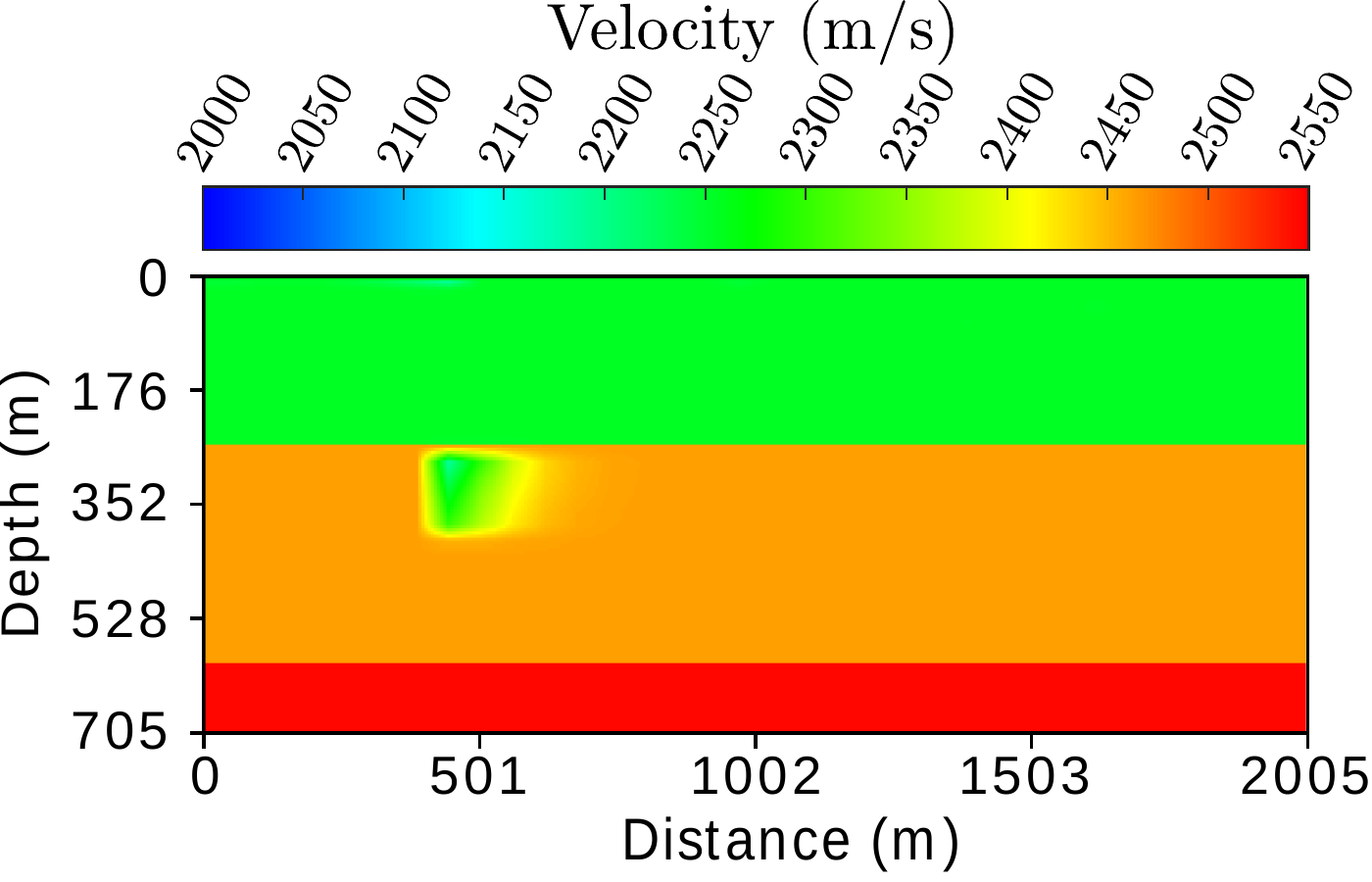}}
      \centerline{\small (a)} 
  \end{minipage}
    \begin{minipage}[b]{0.195\textwidth}
      \centering
      \centerline{\includegraphics[width=1\textwidth]{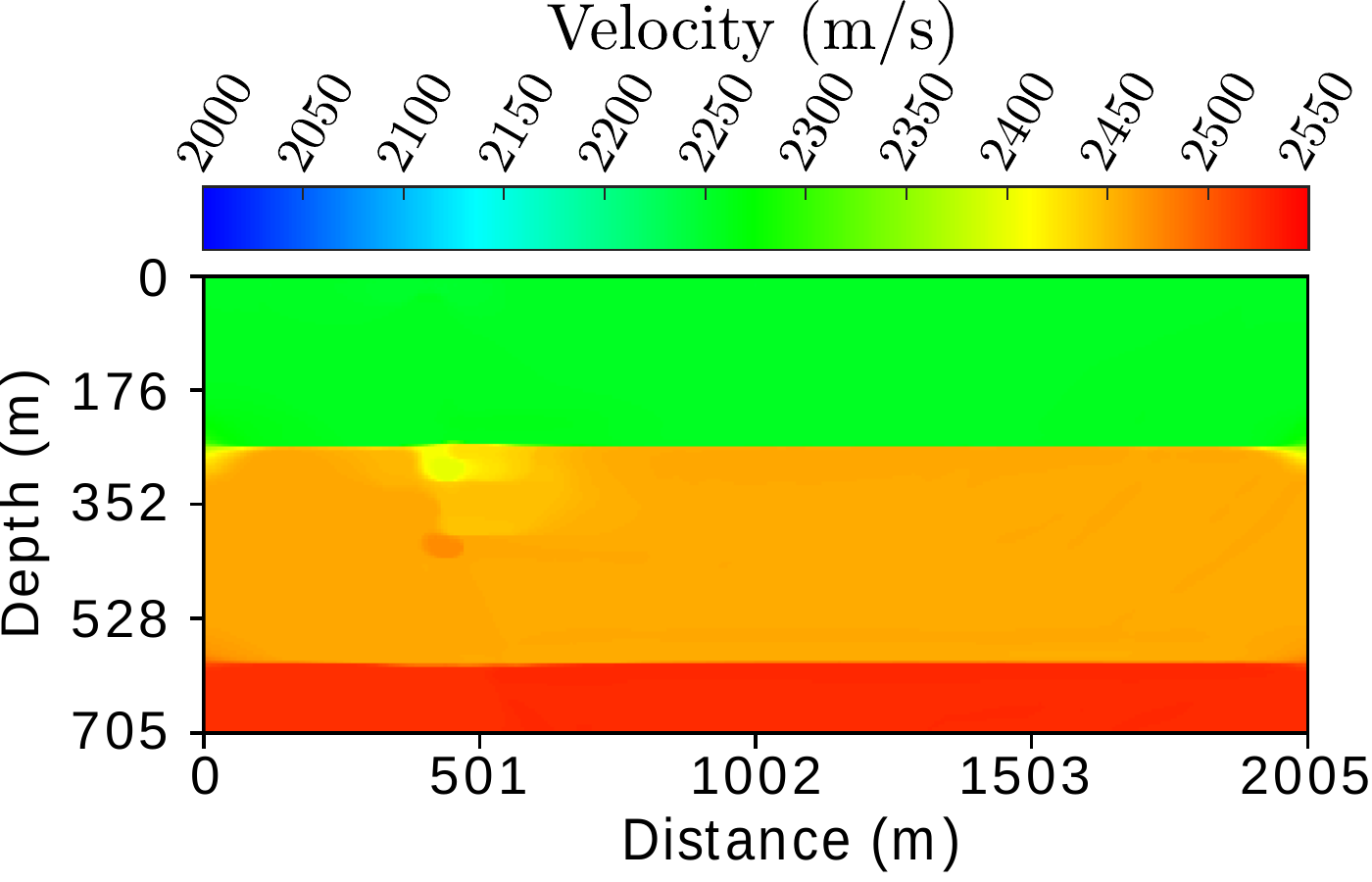}}
      \centerline{\small (b)} 
  \end{minipage}
    \begin{minipage}[b]{0.195\textwidth}
      \centering
      \centerline{\includegraphics[width=1\textwidth]{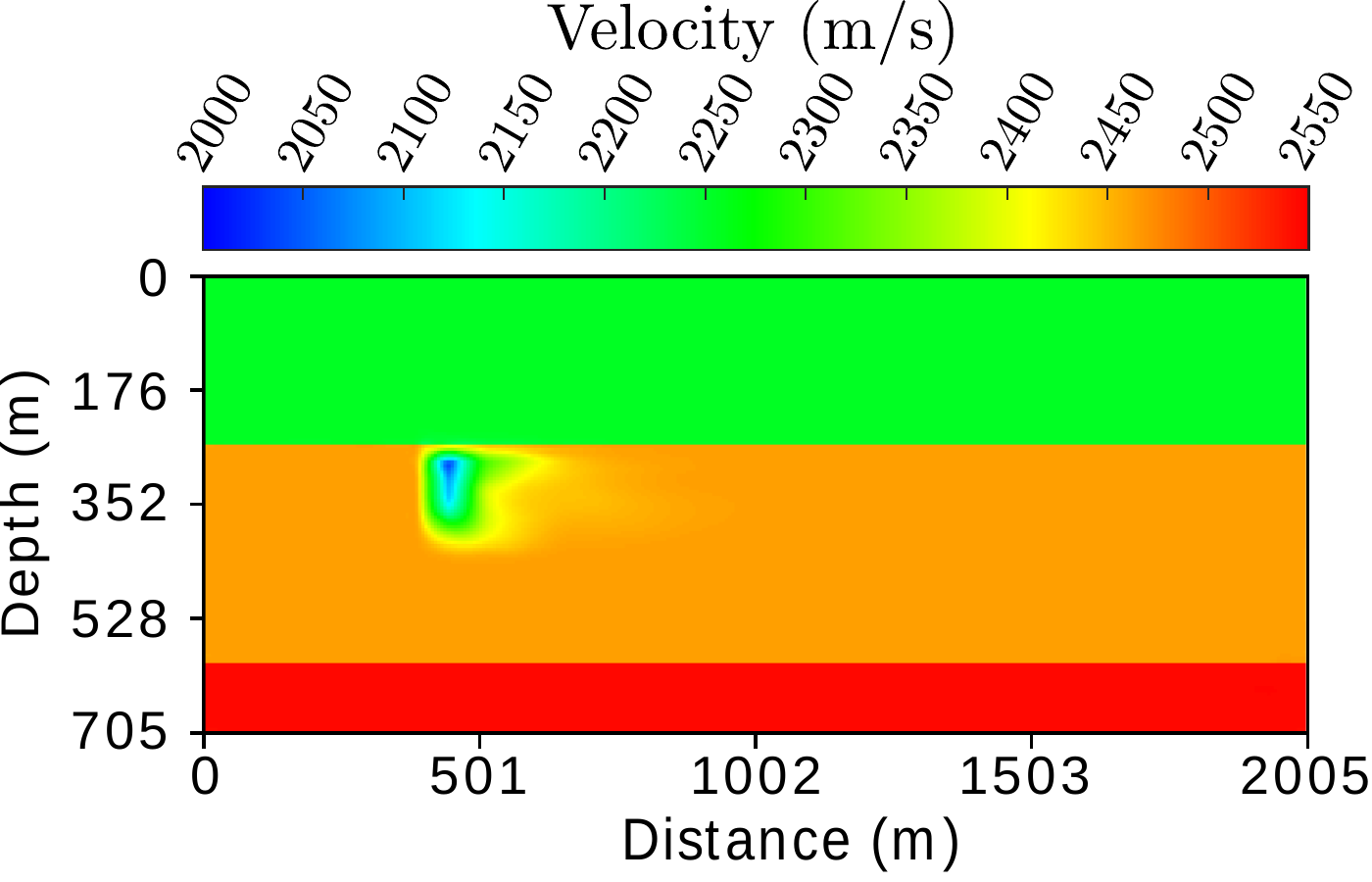}}
      \centerline{\small (c) }
  \end{minipage}
    \begin{minipage}[b]{0.195\textwidth}
    \centering
    \centerline{\includegraphics[width=1\textwidth]{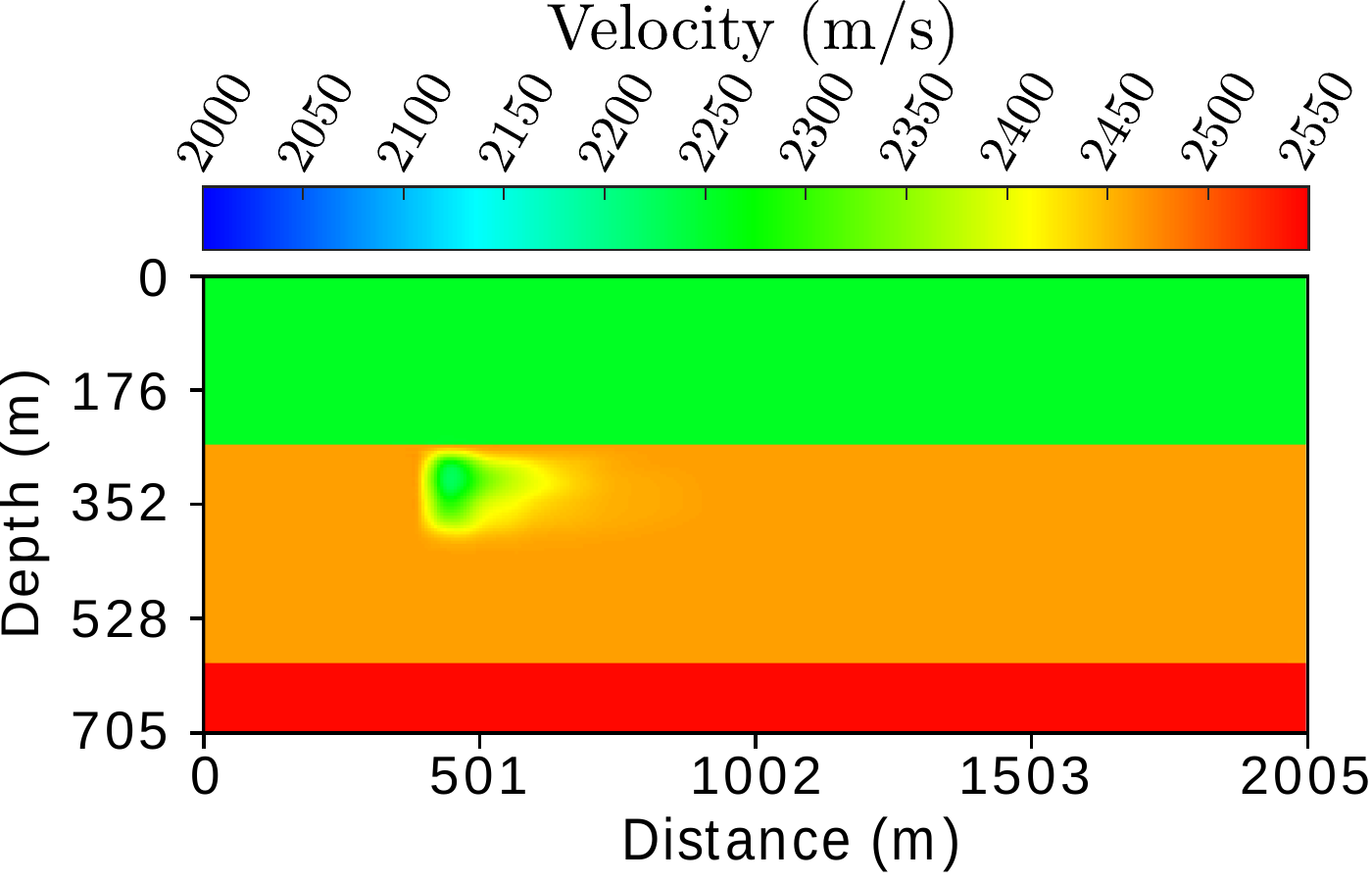}}
    \centerline{\small (d)} 
  \end{minipage}
    \begin{minipage}[b]{0.195\textwidth}
      \centering
      \centerline{\includegraphics[width=1\textwidth]{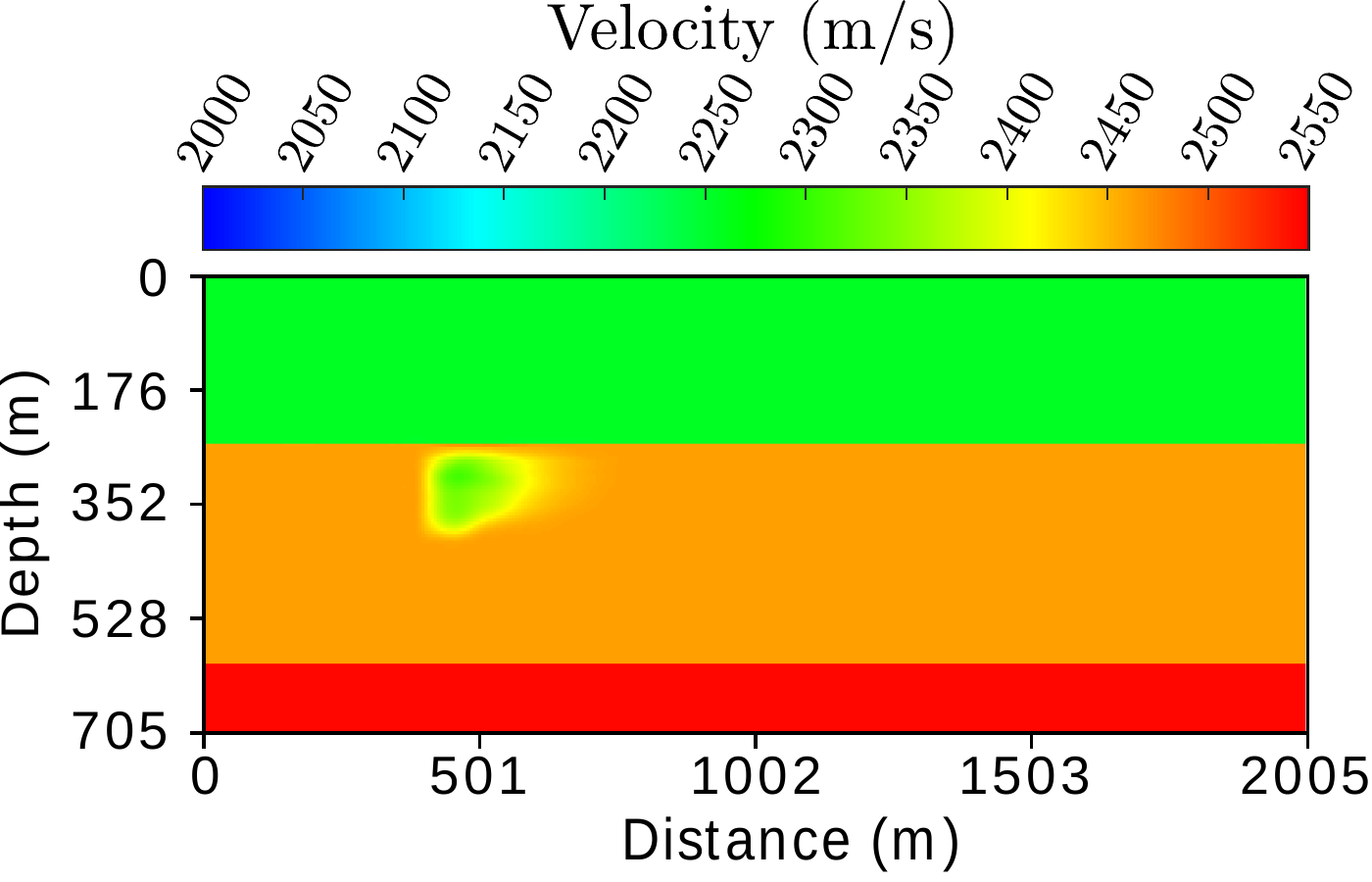}}
      \centerline{\small (e)}
  \end{minipage}  
  }
  \caption{(a) Ground truth. Inversions and errors obtained using (b) physics-based FWI~(0.069, 0.972), data-driven model trained on (c) the \textit{large} and \textit{medium} subsets~(0.0620, 0.990),  (d) \textit{large} and \textit{augmented medium} subsets~(augmented once)~(0.0134, 0.993), and (e) \textit{large} and \textit{augmented medium} subsets~(augmented twice)~(0.0122, 0.994). Error is in the format of (MAE, SSIM).}
  \label{fig_prelim02b}
  \end{figure*}

\subsection{Generalization Performance with Data Augmentation}
\label{sec_transferability}

We evaluate the reconstruction accuracy of the augmentation method described in Section \ref{sec_augmentation}, and compare to traditional data-driven methods with no augmentation strategies. 

For the first scenario, considering the data partitioning described in Section \ref{sec_augmentation}, we compute the reconstruction accuracy along epochs for the following setups:
\begin{enumerate}[i.]
\item Train on the \textit{large} subset.
\item Train on the \textit{large} and \textit{medium} subsets.
\item Train on the \textit{large}, \textit{medium} and \textit{small} subsets.
\item Train on the \textit{large} and augmented \textit{medium} subset.
\end{enumerate}
In every case, performance is evaluated on the  \textit{tiny} subset.

For setups (i), (ii) and (iii), the main differences are the number of training samples and the size of the plumes included in each training set. Based on the network transferability results shown in Section \ref{sec_transferability}, the network trained over \textit{large}, \textit{medium} and \textit{small} plumes is expected to attain the best reconstruction accuracy on \textit{tiny} plumes. On the other hand, setup (iv), corresponding to the network trained over \textit{large} and augmented \textit{medium} subsets, utilizes the same information as setup (ii). However, setup (ii) also includes \textit{medium} samples. In other words, the difference between setup (ii) and (iv) is that setup (iv) increases the number of medium samples using our augmentation strategy.

Figure \ref{fig_loss_test03} shows the reconstruction accuracy along epochs for the four described setups. Regarding the training scenarios with no augmentation strategies, the Mean Absolute Error along epochs behaves as expected: after 250 epochs, training over large, medium and small plumes attains the best reconstruction value of $-13.5$ for the error metric in Eq.~\ref{eq_maelog}, followed by the network trained over large and medium plumes at $-9.3$, and the network trained over large plumes at $-8.7$. On the other hand, our augmented approach from setup (iv) reaches an accuracy of approximately $-12$, which is smaller than the error obtained by setup (ii). This implies that the augmented medium samples generated using the forward model provide physically-coherent information to the training process, allowing a better velocity map reconstruction.

It is important to remark that the augmentation approach from setup (iv) generates new samples by augmenting labeled data from the \textit{medium} subset. Specifically, we train a surrogate network over the labeled \textit{large} subset, use it to generate new data pairs over the labeled \textit{medium} dataset, and then fully train the network using the \textit{large} subset along with both original and augmented \textit{medium} datasets. Although this experimental setup and its comparison against non-augmented strategies illustrates the advantages of our proposed augmentation approach, a more interesting scenario is the generation of physically-consistent pairs from unlabeled data.

With this in mind, a second scenario is considered. Let the \textit{medium unlabeled} subset correspond exclusively to the \textit{medium} seismic data (the corresponding velocity maps are not taken into account). 
Then, consider the following two experiments:
\begin{enumerate}[i.]
    \item Train on the \textit{large} subset.
    \item Train on the \textit{large} and augmented \textit{medium unlabeled} subset.
    \item Test on the \textit{small} subset.
\end{enumerate}

The augmentation process follows the procedure described in Section \ref{sec_augmentation}. In contrast with our first experiment, however, in which both \textit{medium} and augmented \textit{medium} datasets are added to the \textit{large} dataset, only the augmented \textit{medium} dataset is added to the initial \textit{large} dataset. Figure~\ref{fig_loss_test04} shows reconstruction results along $250$ epochs. The network trained over \textit{large} plumes attain a reconstruction accuracy of approximately $-10.5$. On the other hand, the network trained over \textit{large} and augmented \textit{medium unlabeled} plumes obtain a better reconstruction accuracy of approximately $-11.7$. These results strengthen our observation regarding the information encapsulated in the samples generated by the augmentation process: by including the forward modeling operation in the process, physically-consistent data pairs are generated, allowing a better domain adaptation. The use of unlabeled data shows how our proposed method is not limited to labeled data, which can be difficult to obtain in real applications.

Finally, we provide visualization of the inverted velocity maps using both physics-based FWI and data-driven approaches in Fig.~\ref{fig_prelim02b}. Our network trained over the \textit{large} and \textit{medium} subsets obtains a reasonably accurate estimate of the \textit{tiny} plume. Also, the estimate obtained by our augmentation further refines the plume shape and location. This example shows how our augmentation approach used iteratively improves the network output, which reflects the potential of including the physics-based forward model in learning. An additional test and detailed analysis of our inversion model robustness with respect to variant levels of noisy seismic data is provided in the Supplementary Material.

\section{Conclusion and Future Work}
\label{sec:conclusion}

We develop a physics-consistent data-driven seismic full-waveform inversion method. We design a novel data augmentation strategy that incorporates critical physics information and improves the representability of the training set. We validate its performance to detect small \texttt{CO$_{2}$} leakage. Compared with purely physics-based and purely data-driven inversion methods, our physics-consistent data-driven inversion yields higher accuracy and better generalization. With respect to computational cost, the most expensive component of our approach is in data generation~(including modeling) and training.  The cost of the inference is very low. Considering a single inversion, the overall modeling cost of our method can be more expensive (up to 10 times) than that of physics-based FWI. However, for those applications that require inverting multiple seismic surveys at the same location such as time-lapse monitoring, our approach can be very cost-effective.  

Nonrepeatability is an important issue for time-lapse seismic imaging~\cite{Time-2015-Asnaashari}. Our technique is general, and can be combined with existing time-lapse imaging methods. We will study the robustness of our technique in nonrepeatable scenarios. Different means have been proposed to incorporate physics in solving inverse problems, cyclic consistency being one of those~\cite{Unpaired-2017-Zhu}. We will compare the performance between our techniques and consistent generative methods. The high cost in training may hinder the wide application of our method to a broader class of problems. We believe, however, that there may be room for further improving the efficiency of the training. Another future direction would be to explore data augmentation in the low-data regime. 
\section*{Acknowledgements}

This work was supported by the Center for Space and Earth Science at Los Alamos National Laboratory (LANL) and by the Laboratory Directed Research and Development program of LANL under project number 20200061DR. Renán A. Rojas-Gómez would also like to thank Javier E. Santos and Manish Bhattarai for the insightful discussions. We also thank two anonymous reviewers and the Associate Editor, Dr.~Luis~Gómez, for their constructive suggestions and comments that improved the quality of this work.


\bibliographystyle{bst}
\bibliography{egbib}

\end{document}